\documentclass[conference]{IEEEtran}
\usepackage{amsmath,amsfonts,amssymb}
\usepackage{graphicx}
\usepackage{setspace}
\usepackage{tocloft}
\usepackage{mathtools}
\usepackage{footnote}
\ifCLASSINFOpdf
\else
\fi

\begin{document}
\title{Superpixels Based Marker Tracking Vs. Hue Thresholding In Rodent Biomechanics Application}
\author{\IEEEauthorblockN{Omid Haji Maghsoudi}
\IEEEauthorblockA{Spence Lab., Bioengineering\\
Temple University\\
Philadelphia, PA, USA, 19122\\
Email: o.maghsoudi@temple.edu}
\and
\IEEEauthorblockN{A. Vahedipour Tabrizi and B. Robertson}
\IEEEauthorblockA{Spence Lab., Bioengineering\\
Temple University\\
Philadelphia, PA, USA, 19122}
\and
\IEEEauthorblockN{Andrew Spence$^{1}$}
\IEEEauthorblockA{Spence Lab., Bioengineering\\
Temple University\\
Philadelphia, PA, USA, 19122}}
\maketitle
\footnotetext[1]{This material is based upon work supported by, or in part by, the U.S. Army Research Laboratory and the U. S. Army Research Office under contract/grant number W911NF1410141, proposal \#64929EG, to A. Spence. This work supported by Shriners Hospitals for Children Grant \#85115 to Andrew Spence.}
\begin{abstract}
Examining locomotion has improved our basic understanding of motor control and aided in treating motor impairment. Mice and rats are premier models of human disease and increasingly the model systems of choice for basic neuroscience. High frame rates (250 Hz) are needed to quantify the kinematics of these running rodents. Manual tracking, especially for multiple markers, becomes time-consuming and impossible for large sample sizes. Therefore, the need for automatic segmentation of these markers has grown in recent years. We propose two methods to segment and track these markers: first, using SLIC superpixels segmentation with a tracker based on position, speed, shape, and color information of the segmented region in the previous frame; second, using a thresholding on hue channel following up with the same tracker. The comparison showed that the SLIC superpixels method was superior because the segmentation was more reliable and based on both color and spatial information.
\end{abstract}

\IEEEpeerreviewmaketitle

\section{Introduction}
Understanding how animals, including humans, a move is a grand challenge for modern science that has a direct impact on our health and wellbeing. It is a useful instrument with which to explain the biological world and to treat human and animal disease. It most directly impacts the treatment of musculoskeletal injuries and neurological disorders, can improve prosthetic limb design, and aids in the construction of legged robots \cite{maghsoudi2015novel}.

In addition to the intentional changes in gait made by the animal, it is possible to perturb the animal's movement using internal or external perturbation. A mechanical perturbation (e.g., earthquake) while the animal is running, for example, deflecting the surface during running, or an electrical stimulation applied to the nervous system, or even the application of new genetically targeted techniques, like optogenetics or designer receptors exclusively activated by designer drugs \cite{roth2016dreadds}, are several of the increasingly sophisticated methods with which to apply perturbations that dissect movement control. 

In both research and commercial systems, tracking rodents has frequently relied on shaving fur and then drawing markers on the skin for subsequent tracking from raw video, or on the attachment of retroreflective markers, and the use of optical motion capture systems \cite{maghsoudi2016rodent}. 

To track rodents marker (especially on paws), which can provide the required information for gait analysis, several methods have been proposed, including commercially available systems (Digigait \cite{dorman2014comparison, gadalla2014gait}, Motorater, Noldus Catwalk \cite{huehnchen2013assessment, parvathy2013gait, hamers2001automated}). These systems can be prohibitively expensive, and may only provide information about paws during the stance phase. However, although some computerized methods (simple thresholding, cross-correlation, or template matching) has been proposed to answer this need; the usual method to track the markers is manual clicking \cite{hedrick2008software}. 

To have a better understanding of animal locomotion, it is necessary to study some markers indicating different points of the body. These markers can demonstrate posture of rodents including roll, pitch, and yaw \cite{migliaccio2011characterization}. To have such markers for further kinematics studies or better understanding of joints movement before and after a surgery or genetically modification on the animal, some markers are attached or painted on the body. 

Segmentation has been considered as an important part of image processing for different applications \cite{maghsoudi2013detection, penjweini2017investigating, mahdi2017detection, Mori04, Li12, alizadeh2014segmentation}. To resolve this need, we present two methods to segment these markers painted on the body based on superpixels method and hue thresholding. The central contribution of this paper is to demonstrate that SLIC superpixels segmentation performs well for tracking landmarks in video sequences of moving subjects.

\section{Methods}
Four side view cameras were used to capture video from a treadmill. 1000 frames were captured at 250 frames per second, giving 4 seconds per trial. The frames were Bayer encoded and we use a debayering function to convert them to RGB color space frames \cite{maghsoudi2016rodent, maghsoudi2017IET}. We converted to the HSV color space because it places all color information in a single channel, as compared to RGB or LAB colors spaces, in which color is encoded in more than one \cite{maghsoudi2016computer, hajimaghsoudi2012automatic}. Therefore, we transfer the frames to HSV color space and then are able to examine only the hue channel for segmentation based on color.

\subsection{Superpixel Segmentation}
Superpixels contract and group uniform pixels in an image and have been widely used in many computer vision applications such as image segmentation \cite{Mori04, Li12}. The main merit of the superpixel approach is to provide a more natural and perceptually meaningful representation of the input image. There is a large amount of literature on automatic superpixel algorithms, for example, normalized cuts \cite{Ren03}, mean shift algorithm \cite{Comaniciu02}, graph-based method \cite{Felzenszwalb04}, Turbopixels \cite{Levinshtein09}, SLIC superpixels \cite{Achanta12}, and optimization-based superpixels \cite{Veksler10}. Here, we use simple linear iterative clustering (SLIC) \cite{Maghsoudi17WCE}.
\begin{figure}[t!p]
\centering
\includegraphics[width=.33\textwidth]{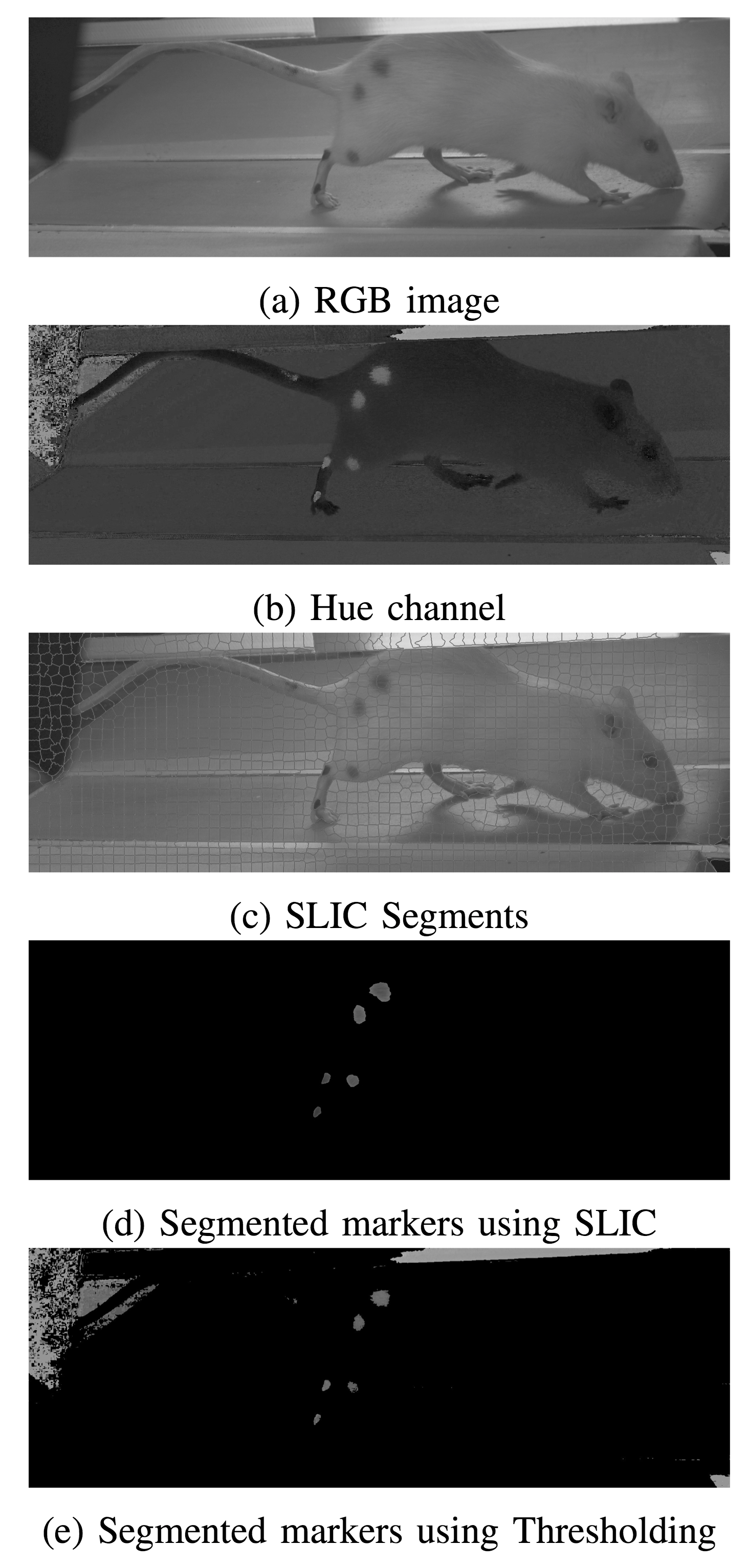}
\caption{A sample video frame of rat locomotion. A, b, c, d, and e show respectively the RGB image, hue channel from HSV color space, the SLIC process on the frame using size 1500 SLIC, the markers segmentation results, and hue threshlding markers segmentation results. Figures are in color (check to the DOI: 10.1109/ACSSC.2017.8335168).} \label{fig1}
\end{figure}
The key parameter for SLIC is size of superpixels. First, $N$ centers are defined as cluster centers. Then, to avoid having centers that are on the edge of an object, the center is transferred to the lowest gradient position in a $3\times3$ neighborhood. The next step is clustering, as each of the pixels are associated with the nearest cluster center based on color information. It means that two coordinate components ($x$ and $y$) depict the location of the segment and three components (for example in the RGB color space, $R$, $G$, and $B$) are derived from color channels. SLIC tries to minimize the residual error and to have this error, a distance (5D Euclidean distance) function is defined as follow:
\begin{equation}
\label{eq:1}
D_{c} = \sqrt{(R_{j}-R_{i})^{2}+(G_{j}-G_{i})^{2}+(B_{j}-B_{i})^{2}},
\end{equation}
\begin{equation}
\label{eq:2}
D_{p} = \sqrt{(x_{j}-x_{i})^{2}+(y_{j}-y_{i})^{2}},
\end{equation}
\begin{equation}
\label{eq:3}
D = \sqrt{(\frac{D_{c}}{N_{c}})^{2}+(\frac{D_{p}}{N_{p}})^{2}}.
\end{equation}
Where $N_{c}$ and $N_{p}$ are respectively maximum distances within a cluster used to normalized the color and spatial proximity. It should be said that SLIC is also constrained to ensure that the region does not grow more than twice the cluster radius; therefore, SLIC size plays an important role on how the segmentation is performed.

SLIC needs to be performed on three or one-dimensional image; therefore, we used RGB color space for SLIC segmentation. The main parameter available for superpixels estimation the superpixels segmented areas is the size of superpixels. Fig 1 shows how the superpixels are performed on an image. 

First, we used a default value for superpixels (1500) based on the average size of markers (200-500 pixels amongst 1,433,600 pixels in a frame). Second, a user was asked to click on the segmented markers; and finally, the superpixels size was updated based on the size of markers in the previous frames to keep the size of superpixels constant compared to the average size of markers (Marker size / Total pixels = Superpixels size). The function updating the size of SLIC superpixels is referred as 'size function'.

\begin{figure*}[t!p]
\centering
\includegraphics[width=0.7\textwidth]{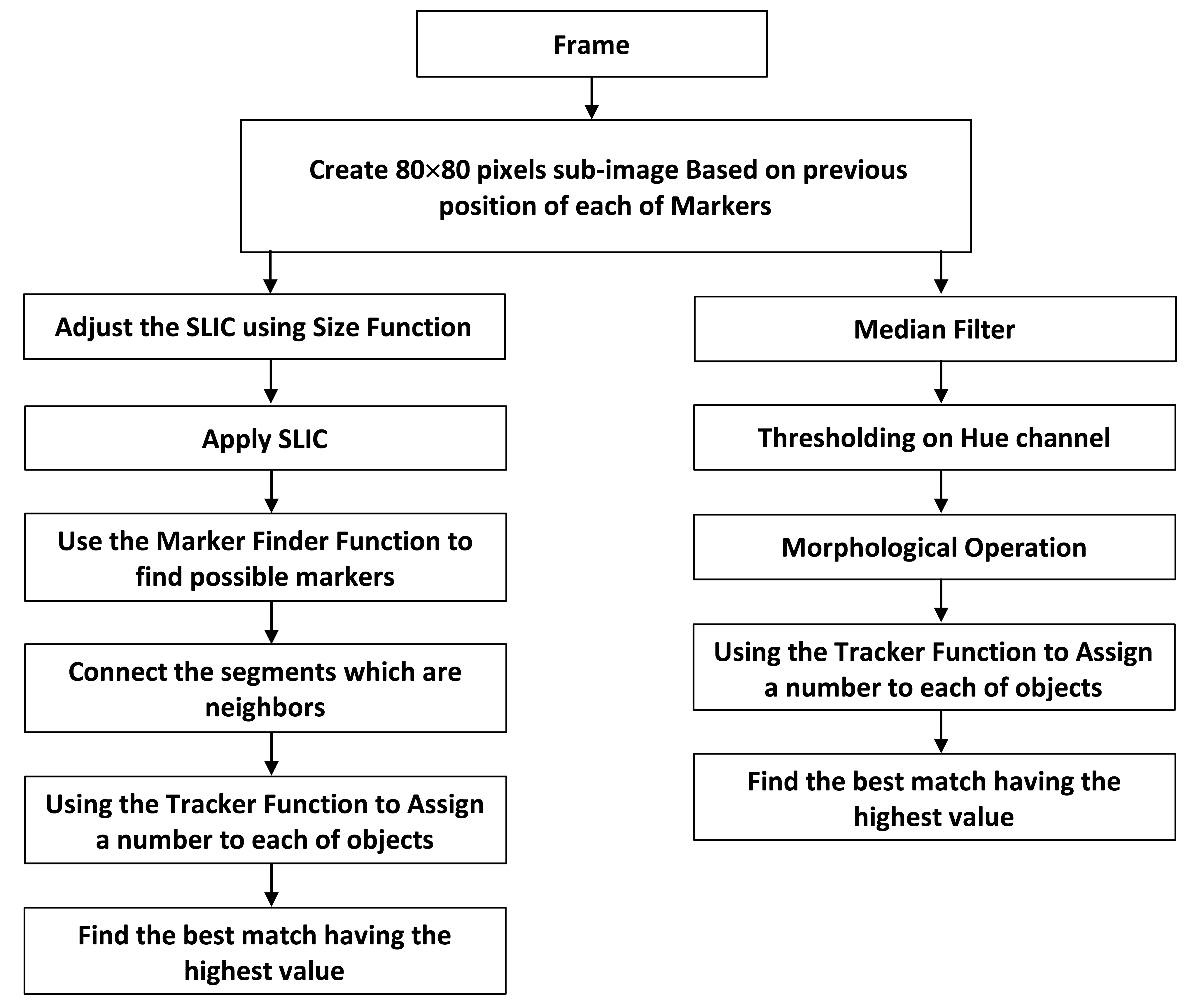}
\caption{Step by step processes to segment and track markers. The left side and right side of chart respectively show the procedure using SLIC superpixels segmentation and thresholding on hue channel. Figures are in color(check to the DOI: 10.1109/ACSSC.2017.8335168).} \label{fig2}
\end{figure*}

We applied a marker finder function to remove the objects having a difference of more than five percent (or ten percent if the marker was not found) between the average of hue value calculated for the previous tracked object and the average of candidate objects in the current frame. It means that if the difference average of hue values between a candidate object in the current frame with the tracked marker in the previous frame was less than five percent (ten percent if the marker was not found) of the hue values average of the marker in the previous frame, it would be considered as a possible marker in the current frame. However, the tracker would find the best match amongst these possible markers. This marker finder function (M) can be formulated as:
\begin{equation}
\tiny C1,I1= \left\{
\begin{array}{   l   l    }
C1 = C1+1, I1 = append(j)   \\
\ \ \ \ \ \ \ \ \ \ \ \ \ (if \ abs(H(SP(j,f)-H(T(i,f-1))\leq 0.05\times H(T(i,f-1))\\ 
\\
C1,I1 \ \  (if \ abs(H(SP(j,f)-H(T(i,f-1))> 0.05\times H(T(i,f-1))\\
\end{array}
\right.
\end{equation} \\
\begin{equation}
\tiny C2,I2= \left\{
\begin{array}{   l   l    }
C2 = C2+1, I2 = append(j)   \\
\ \ \ \ \ \ \ \ \ \  \ \ \ (if \ abs(H(SP(j,f)-H(T(i,f-1))\leq 0.1\times H(T(i,f-1))\\ 
\\
C2,I2 \ \  (if \ abs(H(SP(j,f)-H(T(i,f-1))> 0.1\times H(T(i,f-1))\\
\end{array}
\right.
\end{equation} \\
\begin{equation}
\tiny M(k,i,f)= \left\{
\begin{array}{   l   l    }
Append(SP(n,f)) \ for \  all \ n \ in \ I1\ \ \ \ \ \ \ \  (C1>0)\\
\\
Append(SP(n,f)) \ for \  all \ n \ in \ I2\ \ \ \ \ \ \ \  (C1=0)\\
\end{array}
\right.
\end{equation}
where M, T, H, SP, f, i, and j respectively show marker region (marker finder function), the tracked marker using tracker function for previous frame, hue average value, superpixels region, frame number, number of marker, and number of superpixels. C1 and C2 started by zero in the beginning of procedure. All the objects having borders with each other were connected and considered as one object; it is shown by k in equation 6. 

Finally, another function, size function, was used to update the size of SLIC superpixels. This function estimated the required SLIC size using the size of marker in the previous frame. It was designed based on the fact that SLIC superpixels could be extended up to twice or shrunk to half of the size; therefore, to segment properly, we used the following formula referred as size function:
\begin{multline}
Size \ of \ SLIC \ for \ Frame \ (SSLICF) = \\
\dfrac{2048 (number of columns) \times 700 (number of rows) }{Size \ of \ Marker \ in \ previous \ frame}
\end{multline}
\begin{equation}
Size \ of \ SLIC \ for \ Subimage = \dfrac{SSLICF}{80 \times 80 \times 2}
\end{equation}

\subsection{Hue Segmentation}
As it can be seen in Fig. \ref{fig1}, the marker hue values are completely different with other regions in the frames. To reduce the noise on the frames, it was necessary to use a low pass filter; we chose a median filter with a window size of 10 to smooth the frame. Then, the minimum value of hue was calculated in the marker region and it was deducted by ten percent to provide enough assurance that the threshold will not remove the marker region. The threshold value was 57 (maximum hue value could be 255 in frames) and was selected based on the histogram of markers. To enlarge the circle shape objects (the markers are painted in circle shapes), we used morphological operations; one closing and one opening. The results are illustrated in Fig. \ref{fig1}. 

\subsection{Tracker}
After segmentation using one of the alternate methods, SLIC or thresholding, we use a tracker algorithm that is based on position, speed, size, and color information of the tracked region in the previous frame. After segmentation, a user was asked to click on the right markers on the first frame. We subsequently focused on an $80\times80$ pixel region of interest (ROI) given the user initialization in the first frame, because frame-to-frame marker movement was always within this ROI, and considering only this ROI drastically reduces computation time. The size of the image was selected based on the maximum displacement of the center of the body in rats (30 pixels). Then, we designed a function, tracker function, to assign a weight to each of objects remaining after segmentation. The function found the closest object to the previous tracked marker position, average of hue, size, and following the same speed and direction of movement. The object with the maximum value of this function was chosen as the tracked object in the current frame.

The tracker function can be simplified as follow:
\begin{equation}
\tiny W(k,i,f) = \left\{
\begin{array}{   l   l    }
closest \ object \ to \ T(i,f-1) \\
\ \ \ \ \ \ \ W(k,i,f) = W(k,i,f)+3 \\ 
\\
objects \ moving \ in \ same \ direction \ T(i,f-1) \\ 
\ \ \ \ \ \ \ W(k,i,f) = W(k,i,f)+2 \\
\\
minimum(asb(H(T(i,f-1)- H(M(k,i,f)))) \\
\ \ \ \ \ \ \ W(k,i,f) = W(k,i,f)+2 \\
\\
minimum(asb(S(T(i,f-1)- S(M(k,i,f)))) \\
\ \ \ \ \ \ \ W(k,i,f) = W(k,i,f)+1 \\
\\
minimum(asb(G(T(i,f-1)- G(M(k,i,f)))) \\
\ \ \ \ \ \ \ W(k,i,f) = W(k,i,f)+1
\end{array}
\right.
\end{equation} \\
\begin{equation}
T(i,f) = M(i,f,Maximum(W(k,i,f)))
\end{equation} 
where T and G are respectively the tracked marker for the current frame and average of gray scale image. The whole processes including segmentation and tracking are illustrated in Fig. \ref{fig2}.

\section{Results}
As discussed in the introduction, manual tracking can be considered as the common method to track the markers for many applications in biomechanics. Therefore, we compared the proposed methods with the manually outlined segments. The methods were examined on six Sprague-Dawley rats. Each rat had five markers showing: toe, ankle, knee, hip, and anterior superior iliac spine. We randomly selected two trails from each rat and each trial contains 1,000 frames. It created 12,000 frames captured from one camera. 

\begin{figure}[t]
\centering
\includegraphics[width=0.48\textwidth]{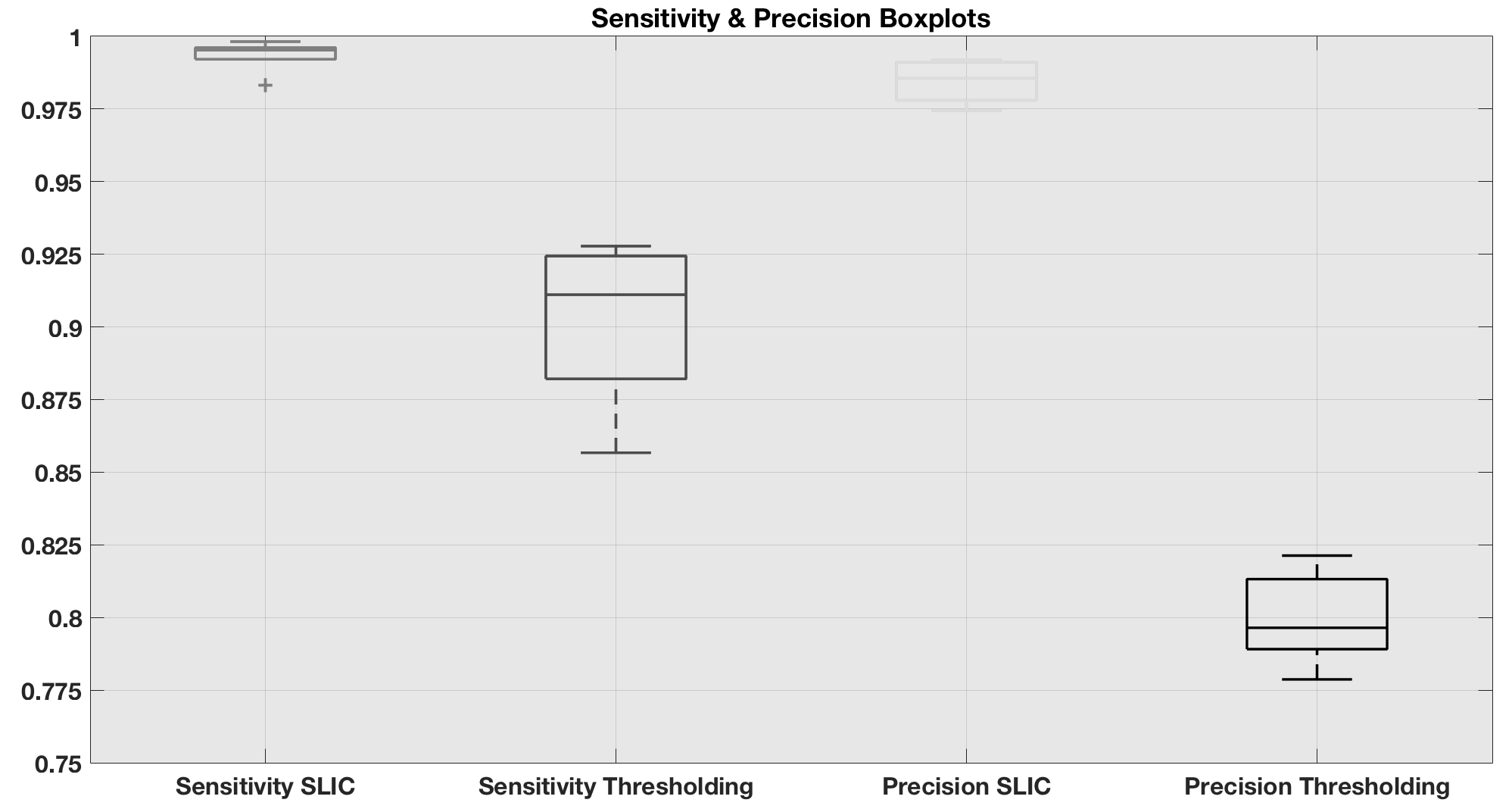}
\caption{Sensitivity and precision boxplots. Red, blue, green, and black boxplots show respectively the boxplots of sensitivity by 
SLIC superpixels, sensitivity by thresholding on hue channel, precision by SLIC superpixels, and precision by thresholding on hue channel. Figures are in color (check to the DOI: 10.1109/ACSSC.2017.8335168).} \label{fig3}
\end{figure}

To evaluate the segmentation process, we randomly selected 15 frames from each trial (totally 180 frames $\times$ 5 markers) and manually outlined the region and compared the results from two methods with the manually outlined regions. Four measures were calculated as follow:
\begin{equation} 
\label{eq:4}
\begin{aligned}
Sensitivity = \dfrac{TP}{TP+FN},\\
Specificity = \dfrac{TN}{TN+FP},\\
Precision = \dfrac{TP}{TP+FP},\\
Accuracy = \dfrac{TP+TN}{TP+TN+FP+FN},
\end{aligned}
\end{equation}
where TP, FP, TN, FN are respectively the number of pixels were segmented by the method and they are matching with the ground truth segmented region, the number of pixels were segmented by the method and they are not matching with the ground truth segmented region, the number of pixels was not segmented by the method and they should not be part of segmentation, and the number of pixels was not segmented by the method and they should be part of segmentation. The results are illustrated in TABLE I. Fig. \ref{fig3} shows the boxplots for the average sensitivity and precision (two measures showed more variation) of the methods.

In addition, to evaluate the tracking performance, we just watched the overlay segmented regions on the frames; there were 574 mistakes (loss of track) for hue thresholding method and just 12 mistakes for SLIC from 60,000 markers (12,000 frames $\times$ 5 markers). 
\begin{table}[b]
\caption{Segmentation results.} 
\label{tab:fonts}
\begin{center}    
\begin{tabular}{ | c | c | c | c | c | c |}
    \hline
    Method & Sensitivity & Specificity & Accuracy & Precision\\ \hline
    SLIC & 0.9933 & 0.9988 & 0.9984 & 0.9845\\ \hline
    Hue Thr & 0.9022 & 0.9829 & 0.9772 & 0.7992\\ \hline
\end{tabular}
\end{center}
\end{table}
\begin{figure*}[t!p]
\centering
\includegraphics[width=0.65\textwidth]{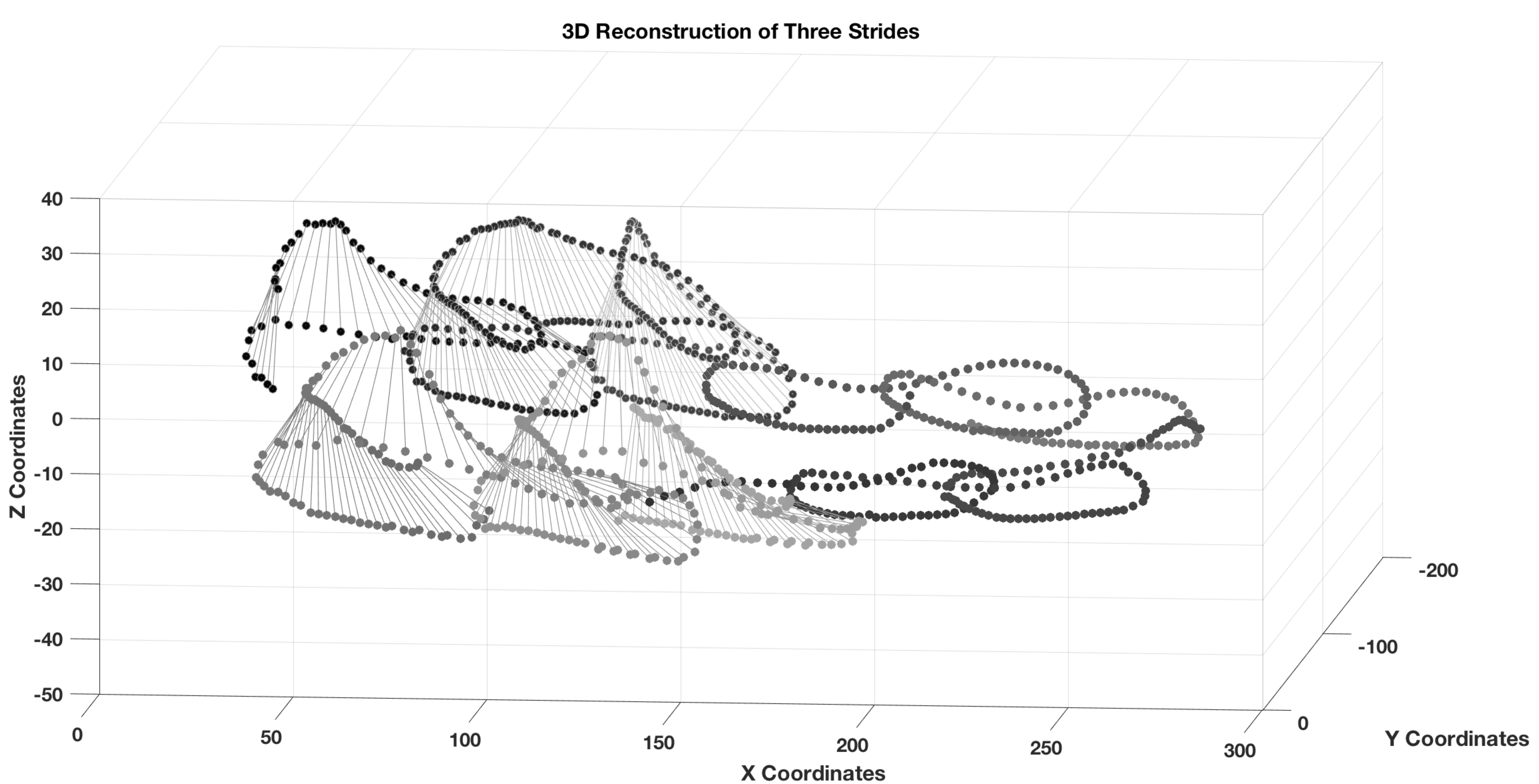}
\caption{3D reconstruction using DLT coefficients for three strides. Figures are in color (check to the DOI: 10.1109/ACSSC.2017.8335168).} \label{fig4}
\end{figure*}
Having these markers tracked in consecutive frames and captured from four cameras provides a promising 3D reconstruction model of markers using direct linear transform (DLT) {\cite{hedrick2008software},\cite{wu2009tracking}}. Achieving an accurate 3D reconstruction will let us to study kinematics in different coordinate systems. Fig. \ref{fig4} shows a 3D reconstructed model of rat with three markers, two on hind paw (showing toe and ankle) and one on front paw.

\section{Conclusion}
Two methods were presented to segment painted markers on the body of rodents. A tracker was designed to find the best-segmented object in the frames based on the position, speed, direction of movement, size, and color information of the tracked marker in the previous frame. As results are illustrated in TABLE I, SLIC superpixel was achieved to a better segmentation rate and led to a better tracking in some cases. The reason came from the point that thresholding needed a low pass filter to smooth the segmented regions and it reduced the size of the markers and it caused disappearing of markers when the markers were so small while SLIC was more accurate for segmentation. These methods are applicable on all rodents as we can select the proper color to provide enough distinctive information from other parts of the body. 

The presented methods play a critical role to analyze big data sets in biomechanics applications. It promises reliable results which can be used for different studies and leads a better 3D reconstruction of the tracked markers as shown in Fig. \ref{fig4}. The latest can provide information about the roll, pitch, and yaw while animal perturbed.

\bibliographystyle{IEEEtran} 
\bibliography{report}

\end{document}